# PERFORMANCE TUNING OF J48 ALGORITHM FOR PREDICTION OF SOIL FERTILITY

Jay Gholap

Dept. of Computer Engineering
College of Engineering, Pune, Maharashtra, India

Abstract:
Data mining involves the systematic analysis of large data sets , and data mining in agricultural soil datasets is exciting and modern research area. The productive capacity of a soil depends on soil fertility. Achieving and maintaining appropriate levels of soil fertility, is of utmost importance if agricultural land is to remain capable of nourishing crop production . In this research, Steps for building a predictive model of soil fertility have been explained.
 This paper aims at predicting soil fertility class using decision tree algorithms in data mining . Further, it focuses on performance tuning of J48 decision tree algorithm with the help of meta-techniques such as attribute selection and boosting.

## 1. INTRODUCTION:

Data mining is a relatively young and interdisciplinary field of computer science, is the process that attempts to discover patterns in large data sets. It utilizes methods at the intersection of artificial intelligence, machine learning, statistics, and database systems. The overall goal of the data mining process is to extract information from a data set and transform it into an understandable structure for further use ("data mining", Wikipedia).

A soil test is the analysis of a soil sample to determine nutrient content, composition and other characteristics. Tests are usually performed to measure fertility and indicate deficiencies that need to be remedied ("Soil Test", Wikipedia)..

In this research , soil dataset containing soil test results has been used to apply various classification techniques in data mining. Soil fertility is a crucial attribute which is considered for land evaluation , also achieving and maintaining necessary levels of fertility is important for nurturing crop production, hence this paper includes steps for building an efficient and accurate predictive model of soil fertility with the help of J48 algorithm.

## 2. RESEARCH METHODOLOGIY

### 2.1. DATASET COLLECTION

Dataset required for this research was collected from private soil testing laboratory in Pune (India) . These datasets contain various attributes and their respective values of soil samples taken from 3 regions of Pune District . Dataset has 10 attributes and a total 1988 instances of soil samples. Table 1 shows attribute description.

| Attribute | Description |
|---|---|
| Ph | pH value of soil |
| EC | Electrical conductivity, decisiemen per meter |
| OC | Organic Carbon, % |
| P | Phosphorous, ppm |
| K | Potassium, ppm |
| Fe | Iron, ppm |
| Zn | Zinc, ppm |
| Mn | Manganese, ppm |
| Cu | Copper, ppm |
| label | Soil fertility class (very low, low, moderate ,moderately high , high, very high) |

Table1: Attribute Description

### 2.2. COMPARISON OF DECISION TREE ALGORITHMS FOR SOIL FERTILITY PREDICTION:

Soil fertility is considered to be one of the critical attributes for deciding cropping pattern in particular area. In this section, results of various decision tree algorithms on dataset are shown. Based on these, the best classifier is selected and further used for tuning its performance. The following section explains decision tree algorithms like J48 , NBTree and SimpleCart in short.

#### 2.2.1. J48 (C4.5):

J48 is an open source Java implementation of the C4.5 algorithm in the Weka data mining tool. C4.5 is a program that creates a decision tree based on a set of labeled input data. This algorithm was developed by Ross Quinlan. The decision trees generated by

C4.5 can be used for classification, and for this reason, C4.5 is often referred to as a statistical classifier ("C4.5 (J48)", Wikipedia).

### 2.2.2. NBTree :

This algorithm is used for generating a decision tree with naive Bayes classifiers at the leaves ( Kohavi R. ,1991) .

### 2.2.3. SimpleCart :

It is a non-parametric decision tree learning technique that produces either classification or regression trees, depending on whether the dependent variable is categorical or numeric, respectively ("CART", Wikipedia).It is used for implementing minimal cost-complexity pruning(Breiman L. et al. 1984)

In this paper, three decision tree techniques ( J48 (C4.5), NBTree and SimpleCart) in data mining were evaluated and compared on basis of accuracy and Error Rate. Tenfold cross-validation was used in the experiment. Our studies showed that J48 (C4.5) model turned out to be best classifier for soil samples.

| Classifier | NBTree | SimpleCart | J48 |
|---|---|---|---|
| Correctly Classified Instances | 1700 | 1824 | 1827 |
| Incorrectly Classified Instances | 288 | 164 | 161 |
| Accuracy (%) | 85.51 | 91.75 | 91.90 |

Table2:
Comparison of different classifiers

### 2.3. TUNING PERFORMANCE OF J48 ALGORITHM

Accuracy of J48 algorithm for predicting soil fertility was highest, hence it was used as a base learner. Now, the aim was to increase its accuracy with the help of some other meta-techniques like attribute selection and boosting with the help of Weka .

### 2.3.1. With attribute selection :

Attribute selection reduces dataset size by removing irrelevant/redundant attributes .It finds minimum set of attributes such that resulting probability distribution of data classes is as close as possible of original distribution. Attribute evaluator method – CfsSubsetEval was used , which evaluates the worth subset of attributes by considering the individual predictive ability of each attribute (Hall M.A. , 1998) . Following are the results using AttributeSelectedClassifier with base learner as J48 .

| Correctly identified instances | 1853 | 93.2093 % |
|---|---|---|
| Incorrectly identified instances | 135 | 6.7907 % |

Table3:
Using AttributeSelectedClassifier with J48 as Base Learner

It can be clearly seen that accuracy has been increased from 91.90 to 93.20 after application of attribute selection technique.

### 2.3.2. Combining attribute selection and boosting method :

Boosting is a machine learning meta-algorithm for performing supervised learning. It can boost performance of weak learner and convert it into a strong learner. It increases the weights of incorrectly identified instances and decreases the weights of correctly identified instances over its iterations.

Adaboost is weka implementation of boosting method which is used for boosting a nominal class classifier (Freund Y. and Schapire R. 1999) . Following are the results after using combination of attribute selection and Adaboost with J48 as base learner.

| Correctly identified instances | 1923 | 96.7304% |
|---|---|---|
| Incorrectly identified instances | 65 | 3.2696% |

Table4:
Results after using combination of attribute selection and boosting with J48 as base learner.

Here, accuracy was enhanced upto 96.73% which makes this predictive model to be more accurate .

### 3. CONCLUSION

The large amounts of data that are nowadays virtually harvested along with the crops have to be analyzed and should be used to their full extent. Various decision tree algorithms can be used for prediction of soil fertility. My studies showed that J48 gives 91.90 % accuracy , hence it can be used as a base learner. With the help of other meta-algorithms like Attribute selection and boosting, J48 gives accuracy of 96.73% which makes a good predictive model .